\documentclass[10pt,a4paper]{article}

\usepackage[T1]{fontenc}
\usepackage[margin=1in]{geometry}
\usepackage{amsmath,amssymb,amsthm}
\usepackage{booktabs}
\usepackage{graphicx}
\usepackage[round]{natbib}
\usepackage[colorlinks=true,citecolor=blue,linkcolor=black,urlcolor=blue]{hyperref}
\usepackage{caption}
\usepackage{xcolor}
\newcommand{\rev}[1]{#1}

\newtheorem{theorem}{Theorem}
\newtheorem{lemma}{Lemma}
\newtheorem{proposition}{Proposition}
\newtheorem{corollary}{Corollary}
\theoremstyle{definition}
\newtheorem{definition}{Definition}

\newcommand{\R}{\mathbb{R}}
\newcommand{\bx}{\mathbf{x}}
\newcommand{\bw}{\mathbf{w}}
\newcommand{\bX}{\mathbf{X}}
\newcommand{\bZ}{\mathbf{Z}}
\newcommand{\bS}{\mathbf{S}}
\newcommand{\cX}{\mathcal{X}}
\newcommand{\cE}{\mathcal{E}}
\newcommand{\cF}{\mathcal{F}}
\newcommand{\cP}{\mathcal{P}}
\newcommand{\Rh}{\widehat{R}}

\newcommand{\rbar}{\bar r}
\newcommand{\rpure}{\bar r^{\,\mathrm{pure}}}
\newcommand{\nn}{\pi}
\newcommand{\Prob}{\mathbb{P}}
\newcommand{\E}{\mathbb{E}}

\newcommand{\AccA}{0.71}
\newcommand{\AccB}{0.87}
\newcommand{\AccC}{0.92}
\newcommand{\AccD}{0.95}

\newcommand{\BudgetLargeFrom}{64}
\newcommand{\BudgetMax}{128}

\newcommand{\CellMin}{1.40}

\newcommand{\GaussBudgetBelow}{32}

\newcommand{\LabelsMax}{148}

\newcommand{\LadderBudget}{128}

\newcommand{\NClassImg}{10}
\newcommand{\NCosRhsBelow}{693}

\newcommand{\NDist}{8}

\newcommand{\NLipAbove}{7}
\newcommand{\NPairs}{10}
\newcommand{\NRunsDet}{6750}
\newcommand{\NRunsPair}{750}

\newcommand{\NZero}{20}

\newcommand{\RepsLadder}{50}
\newcommand{\RepsMain}{50}
\newcommand{\RtCellDig}{195}
\newcommand{\RtCellMn}{2}
\newcommand{\RtCycles}{8, 8, 16, 32 and 64}

\newcommand{\RtHidden}{128}

\newcommand{\RtRbarDeltaMin}{1.05}
\newcommand{\RtRunsPool}{750}
\newcommand{\ShareDigMax}{11.4}
\newcommand{\ShareMnistMax}{7.4}
\newcommand{\ZoRgFeatMax}{0.40}
\newcommand{\ZoRgFeatMin}{0.35}
\newcommand{\TableScaleRows}{digits & 1.22 & 0.12 & 0.33 & 3.3 & 53.9 & 6.9 \\
\addlinespace[2pt]
MNIST: pixels & 3.78 & 0.38 & 1.45 & 18.7 & 89.5 & 27.8 \\
\phantom{MNIST: }pixels, cosine & 0.46 & 0.05 & 0.21 & 8.6 & 88.8 & 20.8 \\
\phantom{MNIST: }PCA (50) & 2.75 & 0.27 & 0.71 & 29.3 & 88.5 & 28.9 \\
\phantom{MNIST: }random features & 1.06 & 0.11 & 0.47 & 40.4 & 93.3 & 35.8 \\
\phantom{MNIST: }features, $10^2$ labels & 3.34 & 0.33 & 0.97 & 28.5 & 87.8 & 28.1 \\
\phantom{MNIST: }features, $10^3$ labels & 8.19 & 0.82 & 2.40 & 40.8 & 90.2 & 32.6 \\
\phantom{MNIST: }features, $5\cdot10^3$ labels & 5.71 & 0.57 & 1.72 & 29.7 & 87.7 & 27.8 \\
\phantom{MNIST: }features, $2\cdot10^4$ labels & 4.04 & 0.40 & 1.33 & 30.1 & 88.0 & 27.7 \\}
\newcommand{\TableBoundRows}{pixels & 0.40 & 0.08 & 0.32 & 3.67 & 0.30 & 3.97 \\
pixels, cosine & 0.79 & 0.76 & 0.03 & 0.79 & 0.25 & 1.04 \\
PCA (50) & 0.35 & 0.11 & 0.24 & 3.49 & 0.26 & 3.75 \\
random features & 0.60 & 0.40 & 0.20 & 1.63 & 0.40 & 2.03 \\
\addlinespace[2pt]
features, $10^2$ labels & 0.28 & 0.08 & 0.20 & 3.60 & 0.29 & 3.89 \\
features, $10^3$ labels & 0.19 & 0.01 & 0.17 & 6.20 & 0.21 & 6.41 \\
features, $5\cdot10^3$ labels & 0.17 & 0.02 & 0.15 & 4.22 & 0.19 & 4.40 \\
features, $2\cdot10^4$ labels & 0.16 & 0.03 & 0.13 & 3.52 & 0.17 & 3.69 \\}
\newcommand{\TableRetrainRows}{Gaussian & $+0.002\pm0.002$ & $-0.003\pm0.003$ & $+0.007\pm0.002$ & $-0.002\pm0.003$ & 0.92 & 0.86 & 0.48 & 0.35 \\
\texttt{digits} & $-0.013\pm0.005$ & $-0.025\pm0.005$ & $-0.031\pm0.004$ & $-0.031\pm0.005$ & 0.00 & 0.00 & 0.04 & 0.00 \\
MNIST & $+0.065\pm0.012$ & $-0.009\pm0.008$ & $-0.044\pm0.004$ & $-0.033\pm0.005$ & 1.00 & 0.54 & 0.02 & 0.15 \\}

\begin{document}

\title{Does coverage explain core-set active learning?\\
Mean, max, and the choice of distance}
\author{Ferdinand Equilbey\\
  {\small Université de Poitiers, Univ. Limoges, CNRS, XLIM, Poitiers, France}\\
  {\small\texttt{ferdinand.equilbey@xlim.fr}}}
\date{November 2026}
\maketitle

\begin{abstract}
Core-set selection casts batch active learning as a covering problem, 
in which the points to label minimise the covering radius of the pool. 
Its main argument is a generalization bound in which this radius controls 
the gap between the loss on the pool and on the labelled sample. 
We examine what this bound requires and what it says in practice. 
We first state it with the assumption that its proof uses at the labelled points, 
a zero conditional risk, and show that it then holds for any dissimilarity 
and any selection rule that uses no label outside the sample. 
The proof works point by point and involves two averages, 
the mean distance of the pool to the labelled sample and the purity of the cells attached to the labelled points. 
Keeping them gives, when labels are deterministic at the labelled points, 
a bound on the pool loss with no zero-loss assumption and without the factor C, 
the number of classes, of the original constant. We then compare scales, 
for deterministic labels and a regression function Lipschitz on the whole pool. 
With the original constant, the bound can be informative only 
if the radius is below the smallest distance between two classes divided by C. 
Without C, its radius form can be informative only once the nearest-neighbour 
rule built on the sample is exact on the pool. 
Finally, a misalignment criterion measures whether a quantity ranks selections 
as the gap or the test error do. Experiments on a Gaussian pool and on two image pools, 
digits and MNIST, the latter under eight distances, and with a network retrained at each cycle, 
show that the radius does not rank selections consistently with the gap and, on the image pools, 
that the bounds that do not use the labels of the pool are never informative 
and that covering in mean rather than in maximum, as greedy k-medoids does, 
gives a higher purity and accuracy than random sampling.

\end{abstract}

\section{Introduction}
\label{sec:intro}

In pool-based batch active learning, a learner repeatedly selects a batch of
unlabelled points, has it labelled by an oracle, and is retrained. The
core-set approach of \citet{sener_active_2018} is the reference
implementation of an exploratory principle: spread the queries so that the
labelled sample covers the support of the data. It selects the batch that
minimises the covering radius of the pool, a $k$-center problem, and
motivates this choice by a generalization bound: the gap between the empirical
risks on the pool and on the labelled sample is at most proportional to that
radius.

A bound can be valid without being an explanation. For it to justify a
selection principle, the quantity it controls must rank selections as the
quantity of interest does. This distinction is familiar for the
generalization of deep networks \citep{jiang_fantastic_2020}. In active
learning, empirical studies report that core-set selection does not
consistently improve on random sampling
\citep{munjal_towards_2022,hacohen_active_2022,yehuda_active_2022,axiotis_data_2024}.
The bound itself has seldom been confronted with the quantities it relates.
\citet{liu_density_2024} report, for one pair of selections, that the one
with the smaller radius has the larger gap; to our knowledge, no systematic
comparison is available. This is the purpose of this paper.

We take the proof at its word. It compares the loss at each pool point with
the loss at its nearest labelled point, and only then replaces every
distance by the largest one. Reading it closely makes its assumptions
explicit, and keeping its averages shows what the argument involves: a mean
distance, and the proportion of pool points whose nearest labelled point has
another label. The second quantity cannot be computed without the labels,
and both change with the distance. Our contributions are the following.
\begin{itemize}
  \item A statement of the bound with explicit assumptions
  (Sect.~\ref{sec:statement}), in which the assumption at the labelled
  points is the zero conditional risk used by its proof, and which holds
  for any dissimilarity and any selection rule that uses no label outside
  the sample.
  \item A reading of the proof (Sect.~\ref{sec:mean}): a deterministic bound
  in terms of the mean distance and the purity, and, when labels are
  deterministic at the labelled points, a bound on the pool loss that
  requires no zero-loss assumption and loses the factor $C$ of the original
  constant.
  \item The scales at which these bounds can be informative
  (Sect.~\ref{sec:scale}), expressed with the smallest distance between two
  classes.
  \item A misalignment criterion (Sect.~\ref{sec:misalignment}) that tests
  whether a quantity optimised by a strategy ranks selections as the
  quantity one hopes to improve.
  \item Experiments (Sect.~\ref{sec:exp}) with three label-free strategies on
  \NPairs{} pool--distance pairs and with a network retrained at each cycle,
  which measure these scales and the
  alignment of the radius, the mean distance and the purity with the gap
  and the test error.
\end{itemize}
We do not propose a selection strategy and we do not question the merits of
$k$-center selection: our purpose is to make explicit what the bound of
\citet{sener_active_2018} requires and to delimit what it allows one to
conclude.

\section{Background}
\label{sec:background}

Let $\cX\subset\R^p$, let $[C]=\{1,\dots,C\}$ be the set of labels, and let
$\bX=\{\bx_1,\dots,\bx_N\}$ be the pool, with labels $y_i$ revealed on demand.
From an initial labelled sample $\bZ_0$ of size $n_0$, a batch of $b$ points
is selected; $\bS\subset[N]$ denotes the labelled sample and $h_{\bS}$ the
model trained on it. A dissimilarity is a nonnegative function $d$ on
$\cX\times\cX$ with $d(\bx,\bx)=0$. We write $\nn(i)\in\bS$ for a labelled
point nearest to $\bx_i$, with $\nn(s)=s$ for $s\in\bS$ and ties broken by
the indices. The covering radius of $\bS$ is
$r_d(\bS)=\max_{i}d(\bx_i,\bx_{\nn(i)})$. Minimising $r_d(\bS)$ over $\bS$
is precisely the $k$-center problem, for which the greedy farthest-first traversal
\citep{gonzalez_clustering_1985} is a $2$-approximation, provided that $d$ is a
metric. We call \emph{core-set} (CS) this greedy selection.

For a given loss $\ell(\bx,c;h)$, let $\Rh_{\bX}(h)$ and $\Rh_{\bS}(h)$ be the
empirical risks of $h$ on the pool and on the labelled sample, and
$\cE(\bS)=|\Rh_{\bX}(h_{\bS})-\Rh_{\bS}(h_{\bS})|$ the \emph{core-set gap}.
\citet{sener_active_2018} bound the population risk of $h_{\bS}$ 
(i.e. the true expectation) by the sum of three terms, 
all of them for the model $h_{\bS}$: the difference between
its population risk and its empirical estimate on the pool, its training loss
$\Rh_{\bS}(h_{\bS})$, and $\cE(\bS)$. The first two being considered small,
the method is designed to control the third, through the bound presented below

\citep[Theorem~1]{sener_active_2018}. It assumes that the pairs $(\bx_i,y_i)_{i\le N}$ are
i.i.d. with $\Prob(y_i=c\mid\bx_i)=:\eta_c(\bx_i)$; for every class $c$, the
loss $\ell(\cdot,c;h)$ is $\lambda^\ell$-Lipschitz and bounded by $L$, and the
regression function $\eta_c$ is $\lambda^\eta$-Lipschitz; the labelled sample
covers the pool with radius $r$; and the model has zero loss at the labelled
points. It then states that, with probability at least $1-\gamma$ over the labels,
\begin{equation}
\label{eq:sener}
  \cE(\bS)\ \le\ r\,\big(\lambda^\ell+\lambda^\eta LC\big)+\sqrt{L^2\log(1/\gamma)/(2N)} .
\end{equation}
The bound gives the method its design principle, which is to minimise $r$.
Section~\ref{sec:statement} states it with explicit assumptions.

Several works revisit this covering argument. \citet{mahmood_low_2022} bound
the difference between the loss on the pool and on a uniformly weighted
sample by the discrete Wasserstein distance between the two, and select the
sample that minimises it. \citet{liu_density_2024} follow the proof of
\eqref{eq:sener} and replace the radius by the largest of the mean distances
within the cells of the labelled points, with the same constant and the same
zero-loss assumption, from which they derive a density-aware variant of the
greedy traversal. \citet{axiotis_data_2024} replace the covering
radius by a $k$-means cost and the uniform sample by an importance-weighted
one. Their guarantee, under a H\"older condition on the loss in an embedding
space, bears on the loss of the current model rather than on that of a model
retrained on the sample; like the cell bound of Sect.~\ref{sec:mean}, it
requires no assumption on the training loss. \citet{yehuda_active_2022} keep balls of fixed
radius and maximise the probability mass that they cover, full coverage
being the high-budget counterpart; their guarantee for the
nearest-neighbour rule rests on the purity of these balls.
\citet{bae_generalized_2024} replace the balls by a smoother notion of
coverage, which a variant of $k$-medoids optimises, and
\citet{li_improving_2022} keep the greedy traversal but scale its distances
by the uncertainty of the model. \citet{hacohen_active_2022} show that
typical rather than atypical examples should be queried when the budget is
small. These works modify the selection principle; we keep the original
bound and ask what it requires and what it measures. Our criterion
transposes to selections the Kendall rank correlations with which
\citet{jiang_fantastic_2020} evaluate complexity measures against the
generalization gap of trained networks. As in their granulated coefficient,
it is restricted to pairs that differ in a single factor, here the strategy.

\section{On the core-set inequality}
\label{sec:theory}

We state the bound with explicit assumptions and keep the averages that its
proof involves. Then we compare the scales at which the resulting bounds can be
informative.

\subsection{Statement and assumptions}
\label{sec:statement}

The proof of \citet{sener_active_2018} compares the expected loss at each
pool point with that at its nearest labelled point, then applies Hoeffding's
inequality to the labels outside the labelled sample. We write it
conditionally on the information available once the sample is labelled,
which makes the role of the selection rule explicit. We say that $\bS$ is
selected \emph{without access to the labels outside $\bS$} if its indices
are chosen one after the other, each as a function of the features
$\bx_{1:N}$, of the labels of the indices already chosen, and of a random
variable $U$ that is independent of the labels given the features. Random
sampling, $k$-center, $k$-medoids and uncertainty-based strategies are of
this kind. We write $y_{\bS}$ for the labels of $\bS$ and
$\cF=\sigma(\bx_{1:N},U,\bS,y_{\bS})$ for the information available once
$\bS$ is labelled; the trained model $h_{\bS}$ is a function of it.

\begin{lemma}
\label{lem:selection}
Let $(\bx_i,y_i)_{i\le N}$ be i.i.d. with
$\Prob(y_i=c\mid\bx_i)=\eta_c(\bx_i)$, and let $\bS$ be selected without
access to the labels outside $\bS$. Given $\cF$, the labels $y_i$,
$i\notin\bS$, are independent, and $y_i$ has distribution $\eta(\bx_i)$.
\end{lemma}

\begin{proof}
Fix $A\subset[N]$. On the event $\{\bS=A\}$ the rule has only read labels of
$A$, and replaying it with these labels returns $A$: this event is a
function of $\bx_{1:N}$, $U$ and $y_A$. Given $\bx_{1:N}$, the labels $y_i$,
$i\notin A$, are independent of $(U,y_A)$ and independent of each other, with
distributions $\eta(\bx_i)$. This remains true given, in addition, $U$,
$y_A$ and an event that is a function of $(\bx_{1:N},U,y_A)$.
\end{proof}

The proof of \citet{sener_active_2018} compares $\bx_i$ with a labelled
point within the covering radius; we take the nearest one, $\bx_{\nn(i)}$.
At this labelled point, its last step uses the zero-loss assumption in the
form of a zero conditional risk, that is, a zero loss for every class of
positive probability. This is assumption (H4) below.

\begin{theorem}[Radius bound, after \citealt{sener_active_2018}]
\label{thm:radius}
Let $(\bx_i,y_i)_{i\le N}$ be i.i.d. with
$\Prob(y_i=c\mid\bx_i)=\eta_c(\bx_i)$, let $\bS$ be selected without access to
the labels outside $\bS$, and let $d$ be a dissimilarity. Write
$d_i=d(\bx_i,\bx_{\nn(i)})$ and consider the following assumptions.
\begin{itemize}
  \item[(H1)] $0\le\ell(\bx_i,c;h_{\bS})\le L$ for every $i\le N$ and $c\in[C]$.
  \item[(H2)] $|\ell(\bx_i,c;h_{\bS})-\ell(\bx_{\nn(i)},c;h_{\bS})|\le\lambda^\ell d_i$ for every $i\le N$ and $c\in[C]$.
  \item[(H3)] $|\eta_c(\bx_i)-\eta_c(\bx_{\nn(i)})|\le\lambda^\eta d_i$ for every $i\le N$ and $c\in[C]$.
  \item[(H4)] $\sum_{c=1}^{C}\eta_c(\bx_s)\,\ell(\bx_s,c;h_{\bS})=0$ for every $s\in\bS$.
\end{itemize}
Let $a_\gamma=L\sqrt{\log(1/\gamma)/(2N)}$. On the event where
\textup{(H1)--(H4)} hold, which belongs to $\cF$,
\begin{equation}
\label{eq:radius}
  \Prob\Big(\cE(\bS)\le r_d(\bS)\big(\lambda^\ell+\lambda^\eta LC\big)+a_\gamma\ \Big|\ \cF\Big)\ \ge\ 1-\gamma .
\end{equation}
\end{theorem}

\begin{proof}
We follow the steps of the proof of \citet[Theorem~1]{sener_active_2018},
with the conditioning on $\cF$, the distances $d_i$ and assumption (H4) made
explicit. Given $\cF$, the sample $\bS$, the model $h_{\bS}$ and the map
$\nn$ are fixed. By (H4) and (H1), $\ell(\bx_s,c;h_{\bS})=0$ for every $s\in\bS$ and
every class $c$ with $\eta_c(\bx_s)>0$. Since $\eta_{y_s}(\bx_s)>0$ almost
surely, the losses of the labelled points vanish: $\Rh_{\bS}(h_{\bS})=0$ and
$\cE(\bS)=\Rh_{\bX}(h_{\bS})=\frac1N\sum_{i\notin\bS}\ell(\bx_i,y_i;h_{\bS})$.
By Lemma~\ref{lem:selection}, for $i\notin\bS$,
\begin{align*}
\E\big[\ell(\bx_i,y_i;h_{\bS})\mid\cF\big]
&=\sum_c\eta_c(\bx_i)\,\ell(\bx_i,c;h_{\bS})\\
&\le\sum_c\eta_c(\bx_{\nn(i)})\,\ell(\bx_i,c;h_{\bS})+\lambda^\eta d_i\sum_c\ell(\bx_i,c;h_{\bS})\\
&\le\sum_c\eta_c(\bx_{\nn(i)})\big[\ell(\bx_{\nn(i)},c;h_{\bS})+\lambda^\ell d_i\big]+\lambda^\eta LC\,d_i\\
&=\lambda^\ell d_i+\lambda^\eta LC\,d_i ,
\end{align*}
where the second line uses (H3) and the nonnegativity of the loss, the
third (H2) and (H1), the latter through
$\sum_c\ell(\bx_i,c;h_{\bS})\le LC$, which is where the factor $C$ enters,
and the last (H4) and $\sum_c\eta_c=1$. As
$d_i\le r_d(\bS)$, summing over $i\notin\bS$ and dividing by $N$ gives
$\E[\Rh_{\bX}(h_{\bS})\mid\cF]\le r_d(\bS)(\lambda^\ell+\lambda^\eta LC)$.
Given $\cF$, the $n=N-|\bS|$ losses $\ell(\bx_i,y_i;h_{\bS})$, $i\notin\bS$,
are independent with values in $[0,L]$. Hoeffding's inequality gives, for
$\varepsilon>0$,
$\Prob(\Rh_{\bX}(h_{\bS})-\E[\Rh_{\bX}(h_{\bS})\mid\cF]\ge\varepsilon\mid\cF)
\le\exp(-2N^2\varepsilon^2/(nL^2))\le\exp(-2N\varepsilon^2/L^2)$,
and $\varepsilon=a_\gamma$ concludes.
\end{proof}

The statement is about a pair $(\bS,d)$, not about $k$-center: the proof uses
neither the optimality of $\bS$ nor the triangle inequality, and it only
needs the Lipschitz properties between a point and its nearest labelled
point. Since it works given $\cF$, the dissimilarity may itself be a
function of $\cF$, such as the distance in a representation learned on the
labelled sample. A change of distance moves the radius and the two Lipschitz constants
together, so that distances must be compared through scale-free quantities
(Sect.~\ref{sec:scale}). Assumption (H4) implies a zero training loss, as
the first step of the proof shows, and coincides with it when labels are
deterministic at the labelled points. When they are random,
it is stronger, and Appendix~\ref{app:h4} shows that a zero training loss
alone does not then suffice for \eqref{eq:sener}. For a loss that vanishes
for at most one class at a given point, such as the $0$--$1$ loss or
$1-p_h(y\mid\bx)$, (H4) holds only if the label of every labelled point is
deterministic, $\eta_{y_s}(\bx_s)=1$, and its loss is zero. Finally, the
constant can be improved: since $\sum_c(\eta_c(\bx_i)-\eta_c(\bx_{\nn(i)}))=0$,
the positive parts of these differences sum to half their absolute values,
and the same proof gives $\lambda^\ell+\tfrac12\lambda^\eta LC$. The next
section removes the factor $C$ altogether when labels are deterministic at
the labelled points.

\subsection{Mean distance and purity}
\label{sec:mean}

The proof compares each pool point with its nearest labelled point, then
bounds every distance by $r$. Three averages appear if this last step is
omitted,
\begin{equation}
\label{eq:rbar-tau}
  \rbar_d(\bS)=\frac1N\sum_{i=1}^N d_i,
  \qquad
  \rpure_d(\bS)=\frac1N\sum_{i:\,y_i=y_{\nn(i)}}d_i,
  \qquad
  \tau_d(\bS)=\frac1N\,\big|\{i:\ y_i\neq y_{\nn(i)}\}\big| ,
\end{equation}
the \emph{mean distance} of the pool to the labelled sample, its restriction
to the pure points, and the \emph{impurity} of the cells $\{i:\nn(i)=s\}$,
$s\in\bS$, which is also the error on the pool of the nearest-neighbour rule
built on $\bS$. Let
$\Rh^{\nn}_{\bS}(h)=\frac1N\sum_i\ell(\bx_{\nn(i)},y_{\nn(i)};h)$ be the loss
on the labelled sample, each labelled point being weighted by the size of
its cell.

\begin{proposition}[Cell bound]
\label{prop:pointwise}
Let $h$ be a model whose loss takes its values in $[0,L]$ and satisfies
$|\ell(\bx_i,c;h)-\ell(\bx_{\nn(i)},c;h)|\le\lambda^\ell d_i$ for every $i$
and $c$. Then
\begin{equation}
\label{eq:pointwise}
  \big|\Rh_{\bX}(h)-\Rh^{\nn}_{\bS}(h)\big|\ \le\ \lambda^\ell\,\rpure_d(\bS)+L\,\tau_d(\bS).
\end{equation}
\end{proposition}

\begin{proof}
If $y_i=y_{\nn(i)}$, then
$|\ell(\bx_i,y_i;h)-\ell(\bx_{\nn(i)},y_{\nn(i)};h)|\le\lambda^\ell d_i$;
otherwise both losses lie in $[0,L]$ and their difference is at most $L$.
Averaging over $i$ gives \eqref{eq:pointwise}.
\end{proof}

Inequality \eqref{eq:pointwise} is deterministic: it needs no assumption on
the sampling of the pool, on $\eta$ or on the selection, no zero-loss
assumption and no confidence term. Its second term requires the labels of
the pool. The Lipschitz property of $\eta$ is what replaces it by a
distance.

\begin{proposition}[Mean-distance bound without the factor $C$]
\label{prop:cfree}
Under the setting of Theorem~\ref{thm:radius}, assume \textup{(H1)--(H3)}
and that labels are deterministic at the labelled points, that is,
$\max_c\eta_c(\bx_s)=1$ for every $s\in\bS$. On the event where these
assumptions hold,
\begin{equation}
\label{eq:cfree}
  \Prob\Big(\Rh_{\bX}(h_{\bS})\le\Rh^{\nn}_{\bS}(h_{\bS})+\big(\lambda^\ell+L\lambda^\eta\big)\,\rbar_d(\bS)+a_\gamma\ \Big|\ \cF\Big)\ \ge\ 1-\gamma .
\end{equation}
If labels are deterministic at every pool point, the inequality inside the
probability holds almost surely with $a_\gamma=0$.
\end{proposition}

\begin{proof}
Let $i\notin\bS$ and $j=\nn(i)$. By Lemma~\ref{lem:selection}, bounding by $L$
the loss of the classes other than $y_j$,
\begin{align*}
\E\big[\ell(\bx_i,y_i;h_{\bS})\mid\cF\big]
=\sum_c\eta_c(\bx_i)\,\ell(\bx_i,c;h_{\bS})
&\le\ell(\bx_i,y_j;h_{\bS})+L\,\big(1-\eta_{y_j}(\bx_i)\big)\\
&\le\ell(\bx_j,y_j;h_{\bS})+\lambda^\ell d_i+L\lambda^\eta d_i ,
\end{align*}
where the last step uses (H2) and
$1-\eta_{y_j}(\bx_i)=\eta_{y_j}(\bx_j)-\eta_{y_j}(\bx_i)\le\lambda^\eta d_i$,
since $\eta_{y_j}(\bx_j)=1$ almost surely. 
For $i\in\bS$, $\nn(i)=i$ and $d_i=0$. Summing over $i$ and dividing by $N$,
$\E[\Rh_{\bX}(h_{\bS})\mid\cF]\le\Rh^{\nn}_{\bS}(h_{\bS})+(\lambda^\ell+L\lambda^\eta)\rbar_d(\bS)$,
and Hoeffding's inequality concludes as in the proof of
Theorem~\ref{thm:radius}. If labels are deterministic at every pool point,
the conditional distribution of the labels given $\cF$ is degenerate and
$\Rh_{\bX}(h_{\bS})$ equals its conditional expectation.
\end{proof}

Proposition~\ref{prop:cfree} differs from \eqref{eq:radius} in three
respects. The constant has lost its factor $C$: the full loss $L$ is charged
only on the event that the label differs from that of the nearest labelled
point, whose probability is at most $\lambda^\eta d_i$. The distance is a
mean and not a maximum; with $\rbar_d\le r_d$ the same bound holds with the
radius. The training loss is not assumed to vanish: it appears on the
right-hand side, weighted by the cells, and when it is zero the left-hand
side is the gap $\cE(\bS)$. The price is the assumption on the labels of the
labelled points, which (H4) already implies for the losses mentioned above.
Under deterministic labels, an impure point is at distance at least
$1/\lambda^\eta$ from its labelled point, so that
$\tau_d\le\lambda^\eta\rbar_d$: the right-hand side of \eqref{eq:cfree} is
then at least $\Rh^{\nn}_{\bS}+\lambda^\ell\rpure_d+L\tau_d$, the bound on
the pool loss given by \eqref{eq:pointwise}. The bounds \eqref{eq:sener},
\eqref{eq:radius} and \eqref{eq:cfree} thus rest on two relaxations of the
cell bound: the mean distance is replaced by the largest one, and the
impurity by its worst case over the labelling functions that are
$\lambda^\eta$-Lipschitz.

Neither quantity is new. The mean distance is the objective of $k$-medoids
\citep{kaufman_finding_1990}, and it is at most the Wasserstein distance of
\citet{mahmood_low_2022}, which imposes uniform weights on the sample; the
largest of the cell averages of the distance is the quantity of
\citet{liu_density_2024}; the impurity is close to the purity of
\citet{yehuda_active_2022}. What matters
here is their status: the radius and the mean distance can be minimised
without labels, the impurity cannot.

\subsection{Informative scales}
\label{sec:scale}

Since the loss is bounded by $L$, so are the pool loss and the gap, and a
bound carries information only if its right-hand side is below $L$. We say
that it is then \emph{informative}. Under deterministic labels, this
requirement can be expressed with the geometry of the labelled pool.

\begin{proposition}[Scales]
\label{prop:scale}
Assume that labels are deterministic at every pool point, and let $\Delta_d$
be the smallest distance between two pool points with different labels. Let
$\lambda^\eta$ be a Lipschitz constant of the functions $\eta_c$ on the pool,
so that \textup{(H3)} holds whatever the sample. Then
$\lambda^\eta\ge1/\Delta_d$, and the following holds for every
$\lambda^\ell\ge0$.
\begin{enumerate}
  \item The right-hand side of \eqref{eq:sener} and \eqref{eq:radius} is
  below $L$ if and only if
  $r_d(\bS)<(L-a_\gamma)/(\lambda^\ell+\lambda^\eta LC)$, which requires
  $r_d(\bS)<(1-a_\gamma/L)\,\Delta_d/C$.
  \item The right-hand side of \eqref{eq:cfree} is below $L$ only if
  $\rbar_d(\bS)<(L-a_\gamma)/(\lambda^\ell+L\lambda^\eta)$, which requires
  $\rbar_d(\bS)<\Delta_d$. With the radius in place of the mean distance, it
  requires $r_d(\bS)<\Delta_d$.
  \item If $r_d(\bS)<\Delta_d$, then $\tau_d(\bS)=0$: the nearest-neighbour
  rule built on $\bS$ labels every pool point correctly.
\end{enumerate}
\end{proposition}

\begin{proof}
Two pool points $\bx,\bx'$ with labels $y\neq y'$ at distance $\Delta_d$ give
$1=|\eta_y(\bx)-\eta_y(\bx')|\le\lambda^\eta\Delta_d$. The thresholds follow
from $\lambda^\ell\ge0$, $\Rh^{\nn}_{\bS}\ge0$ and $\lambda^\eta\ge1/\Delta_d$.
For the last claim, a labelled point with another label than $\bx_i$ is at
distance at least $\Delta_d$ from it, hence is not its nearest labelled
point when $r_d(\bS)<\Delta_d$.
\end{proof}

\begin{corollary}
\label{cor:budget}
Let $d$ be a metric, $\hat r_k$ the radius reached by the greedy $k$-center
traversal with $k$ centers, started from an arbitrary point, and $\rho>0$. A
sample with $r_d(\bS)<\rho$ has at least $\min\{k:\hat r_k<2\rho\}$ points.
If $\rho$ is smaller than every nearest-neighbour distance in the pool, it
is the whole pool.
\end{corollary}

\begin{proof}
A sample of $k$ points with radius below $\rho$ shows that the optimal
$k$-center radius is below $\rho$, and the greedy radius is at most twice
the optimal one \citep{gonzalez_clustering_1985}: $\hat r_k<2\rho$. For the
second claim, a point outside $\bS$ is at least as far from $\bS$ as from
its nearest neighbour in the pool.
\end{proof}

The constants of Theorem~\ref{thm:radius} and of Prop.~\ref{prop:cfree}
thus set two scales. With the factor $C$, the bound can be informative only
at a scale $C$ times finer than one at which the nearest-neighbour rule
built on the sample is already exact on the pool. This factor is a property
of the constant and not of the argument; with the constant
$\lambda^\ell+\frac12\lambda^\eta LC$ of Sect.~\ref{sec:statement}, the first
scale is $2\Delta_d/C$. Without it, two statements remain.
In its radius form, the bound can be informative only when the
nearest-neighbour rule built on $\bS$ is exact on the pool, that is, when
the labels of the pool are already known. In its mean form, the condition
$\rbar_d<\Delta_d$ is weaker, since a few isolated points can keep the radius
large while the mean is small; it is necessary and not sufficient, the
exact threshold depending on the model through $\lambda^\ell$. The ratios
$r_d/\Delta_d$ and $\rbar_d/\Delta_d$ do not change when $d$ is multiplied by
a constant, so that they can be compared across distances.
Section~\ref{sec:exp-scale} measures them.

These conditions bear on a constant $\lambda^\eta$ that is valid for every
sample, as in \eqref{eq:sener}. For a given sample, (H3) only involves the
pairs formed by a pool point and its nearest labelled point, and it may hold
with a smaller constant; but that constant depends on the labels of the
pool, and so does not provide a guarantee before they are known.

\section{Misalignment criterion}
\label{sec:misalignment}

A bound $v\le\varphi(u)$ with $\varphi$ increasing suggests to choose,
among candidate selections, the one with the smallest $u$. Whether this
choice also yields a small $v$ is an empirical question about the ranking of
selections, on which the validity of the inequality is silent.

\begin{definition}[Misalignment]
\label{def:mis}
Let $\cP$ be a set of pairs of labelled samples to be compared, and $u,v$
two real functionals of a labelled sample. The misalignment of $u$ and $v$
on $\cP$ is the proportion of pairs that they order oppositely,
\begin{equation}
\label{eq:mis}
  m(u,v)=\frac{1}{|\cP|}\sum_{(\bS,\bS')\in\cP}\Big(\mathbf{1}\{\delta u\,\delta v<0\}+\tfrac12\,\mathbf{1}\{\delta u\,\delta v=0\}\Big),
\end{equation}
with $\delta u=u(\bS)-u(\bS')$ and $\delta v=v(\bS)-v(\bS')$: a pair on which
$u$ or $v$ is tied counts for one half.
\end{definition}

A value $m=0$ means that $u$ ranks the selections as $v$ does, $m=1/2$ that
\rev{the pairs ordered in the same way and in opposite ways balance}, and $m>1/2$ that choosing the smaller
$u$ gives the larger $v$ more often than not; without ties,
$m=(1-\tau_{\mathrm K})/2$ for Kendall's rank correlation $\tau_{\mathrm K}$
restricted to $\cP$. We take for $\cP$ the pairs of selections produced by
two strategies from the same pool, the same initial sample and the same
budget, so that only the strategy differs. For $u$ we consider the three
quantities of Sect.~\ref{sec:theory}: the maximal distance $r_d$, which
$k$-center minimises and the bound states; the mean distance $\rbar_d$ and
the impurity $\tau_d$, which the proof uses. For $v$ we take the gap $\cE$,
which the bound \eqref{eq:sener} is about, and the test error $R$, which the practitioner
cares about. A value close to $1/2$ has two possible meanings: \rev{$u$ does not
order $v$ consistently}, or the selections compared do not differ in $v$ beyond noise.
A value whose interval lies above $1/2$ is neither: it is a systematic
disagreement.

\section{Experiments}
\label{sec:exp}

We measure the scales of Sect.~\ref{sec:scale} on our pools and compare the
selections of three strategies through the quantities of
Sect.~\ref{sec:theory}. Appendix~\ref{app:details} gives the experimental
details, and Appendix~\ref{app:results} the misalignments, the choice of
distance, a network retrained at each cycle and the terms of the cell bound.

\subsection{Protocol}
\label{sec:exp-protocol}

We use two small pools with the Euclidean distance: two Gaussian classes in
the plane with a shared covariance ($N=600$, $C=2$, Bayes error $0.15$), for
which $\eta$ is logistic and $\lambda^\eta$ is known, and the \texttt{digits}
images of \texttt{scikit-learn} \citep{pedregosa_scikit_2011} ($N=1297$,
$p=64$, $C=\NClassImg$). A third pool of $2000$ MNIST images
\citep{lecun_gradient_1998}, with $C=\NClassImg$, is equipped with \NDist{}
distances, each being the Euclidean distance in a representation. Four are
label-blind (raw pixels, pixels normalised to unit norm, which give the
cosine distance, $50$ principal components and $128$ random ReLU features),
and four are learned by a network on $10^2$ to $2\cdot10^4$ labelled images
disjoint from the pool. Labels are random on the Gaussian pool, and we treat
them as deterministic on the two image pools, where every image has one
label.

One repetition draws $n_0=\NZero$ initial points and selects
$b\in\{8,16,32,64,128\}$ further points with three label-free strategies:
uniform sampling, CS, and greedy $k$-medoids (KM), which adds at each step
the point that most reduces $\rbar_d$. A logistic regression $h_{\bS}$ is
fitted on the labelled points, in the representation that defines the
distance. Its loss $\ell(\bx,y;h)=1-p_h(y\mid\bx)$ takes its values in
$[0,1]$, so that $L=1$, and it is Lipschitz in $\bx$ with constant
$\lambda^\ell=\frac14\max_{a,b}\|\bw_a-\bw_b\|$, where $\bw_a$ is the weight
vector of class $a$. We record $r_d$, $\rbar_d$, $\tau_d$, the gap $\cE$ for
this loss and the test error, over \RepsMain{} repetitions.

\subsection{Measured scales}
\label{sec:exp-scale}

\begin{table}[t]
  \centering
  \caption{Pools with deterministic labels: scales of the bounds.
  $\Delta_d$: smallest distance between two classes; $\Delta_d/C$: scale of
  the constant of \eqref{eq:sener}; $d_{\mathrm{nn}}$: smallest
  nearest-neighbour distance in the pool. Last three columns, in percent of
  the pool: lower bound on the size of any sample with $r_d<\Delta_d$
  (Corollary~\ref{cor:budget}); labels used by greedy $k$-center when its
  radius falls below $\Delta_d$; labels used by greedy $k$-medoids when its
  mean distance falls below $\Delta_d$. Our largest samples, of \LabelsMax{}
  labels, are \ShareDigMax\,\% of \texttt{digits} and \ShareMnistMax\,\% of
  the MNIST pool.}
  \label{tab:scale}
  \small\setlength{\tabcolsep}{5pt}
  \begin{tabular}{lrrrrrr}
    \toprule
     & & & & \multicolumn{2}{c}{$r_d<\Delta_d$} & $\rbar_d<\Delta_d$ \\
    \cmidrule(lr){5-6}\cmidrule(lr){7-7}
    Pool and distance & $\Delta_d$ & $\Delta_d/C$ & $d_{\mathrm{nn}}$ & any $\bS$ ($\ge$) & $k$-center & $k$-medoids \\
    \midrule
    \TableScaleRows
    \bottomrule
  \end{tabular}
\end{table}

\rev{For both image pools and every fixed distance}, $\Delta_d/C$, and even $2\Delta_d/C$,
is below the smallest nearest-neighbour distance of the pool
(Table~\ref{tab:scale}), so that, by Corollary~\ref{cor:budget}, the scale
of the constant of \eqref{eq:sener} is reached only when the whole pool is
labelled. Without the factor $C$, the radius form requires $r_d<\Delta_d$,
out of reach of any sample of \LabelsMax{} labels on the MNIST pool and
never reached on \texttt{digits}, and the mean form requires
$\rbar_d<\Delta_d$, reached only by $k$-medoids on \texttt{digits} at
$b=\BudgetMax$, where the Lipschitz term of the fitted model exceeds $L$.
With this Lipschitz constant and \rev{any Lipschitz constant of $\eta$ on
the pool}, none of the bounds on the pool loss is informative in any run on
the image pools \rev{with fixed representations}
(Appendix~\ref{sec:exp-terms}).

The assumptions of the bounds fail as well. The loss $1-p_h$ never
vanishes, so that (H4) holds on none of our pools, and the gap $\cE$, which
\eqref{eq:sener} is about, is covered by none of the bounds. On the Gaussian
pool, the scale of \eqref{eq:sener} is reached from $b=\GaussBudgetBelow$ on,
but neither (H4) nor the assumption of Prop.~\ref{prop:cfree} can hold, both
classes having a positive probability everywhere.

\subsection{Maximum, mean and purity}
\label{sec:exp-align}

\begin{figure}[t]
  \centering
  \includegraphics[width=\textwidth]{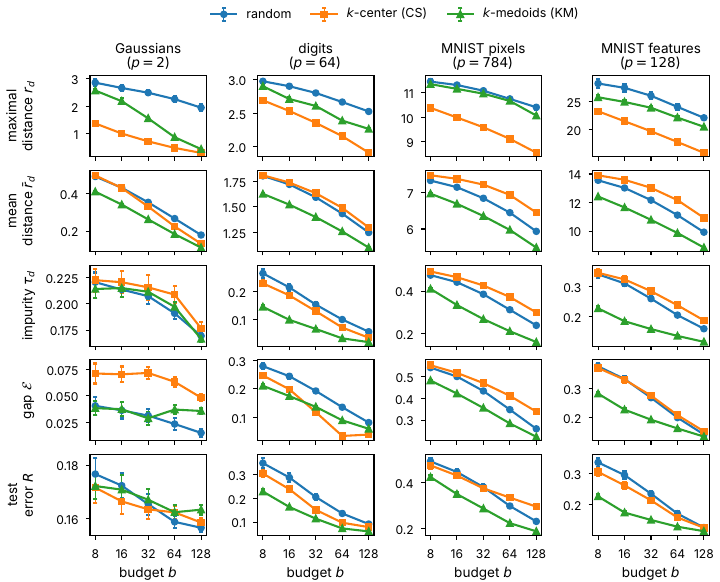}
  \caption{The quantities of Sect.~\ref{sec:theory}, the core-set gap of
  the loss $1-p_h(y\mid\bx)$ and the test error as functions of the
  budget, on four pool--distance pairs (Gaussian pool, \texttt{digits},
  MNIST pixels and MNIST features learned on $5\cdot10^3$ labels; mean and
  $95\,\%$ interval over \RepsMain{} repetitions).}
  \label{fig:quantities}
\end{figure}

$k$-center has the smallest radius, as designed, but not the smallest
mean distance (Fig.~\ref{fig:quantities}). On the MNIST pool its mean
distance even exceeds that of random sampling, since lowering the maximum
amounts to labelling isolated points, which leaves the bulk of the pool
farther from its nearest labelled point; \citet{axiotis_data_2024} also note
that the $k$-center objective is sensitive to outliers. $k$-medoids, which
minimises the mean distance, has the lowest impurity and the lowest test
error on the three image pairs. On the Gaussian pool, where the classes
overlap, the three strategies differ little in impurity and in test error.

The gap does not follow the radius. In the terms of
Definition~\ref{def:mis}, the radius is opposed to the gap on the Gaussian
pool and on MNIST pixels and aligned with it on \texttt{digits} at the
larger budgets, whereas the mean distance and the impurity are aligned with
it on the two MNIST pairs (Fig.~\ref{fig:misalignment}). On the Gaussian
pool and on \texttt{digits}, the differences of gap between CS and random
sampling come from the training loss, the losses on the pool being nearly
equal, so that the gap says how representative the training loss is rather
than how good the model is.

Appendices~\ref{sec:exp-dist} and~\ref{sec:exp-retrain} vary the
distance of the MNIST pool and retrain the network at each cycle. CS is less
accurate than random sampling under the label-blind distances, but more
accurate on the image pools with the retrained network, under which \rev{no
consistent ordering of the gap by the radius is detected} on MNIST. On the image pools,
$k$-medoids has a lower impurity and a lower test error than random sampling
in all these settings.

\section{Discussion and limitations}
\label{sec:discussion}

The bound describes a regime, in which labels are deterministic at the
labelled points, the model has zero loss there, and the classes are
separated at a scale that the labelled sample resolves. With the constant of
\eqref{eq:sener}, this scale is $C$ times finer than the one at which the
nearest-neighbour rule built on the sample is exact on the pool. Without the
factor $C$, the radius form still requires this rule to be exact, and the
mean form requires a mean distance below the separation of the classes, a
condition that $k$-medoids reaches on \texttt{digits} without the bound
becoming informative. On our image pools, no bound on the pool loss falls
below the trivial one with fixed representations, and only the cell bound,
which requires the labels of the pool, does so with the retrained network,
in a minority of runs.

Outside this regime, which quantity accounts for the behaviour of a
selection is an empirical question, and the radius does not do so
consistently, whether the representation is fixed or retrained. The gap
itself is an indirect target, since its differences can come from the
training loss alone. The purity is on the scale of what the cell bound
controls. For $k$-center against random sampling on the image pools, it is
aligned with the gap under every fixed distance but the cosine one and with
the retrained network, and over the three strategies it is nearly always
the quantity most aligned with the test error on these pools
(Fig.~\ref{fig:misalignment}).
But it requires the labels, and the mean distance, its label-free
relaxation, does not rank the gap consistently. A label-free strategy can
nevertheless act on the purity through the geometry. On the image pools,
covering in mean rather than in maximum, as greedy $k$-medoids does,
improves the purity and the accuracy over random sampling under every
distance that we tried, the retrained one included, in line with the
low-budget literature
\citep{hacohen_active_2022,yehuda_active_2022,bae_generalized_2024} and with
clustering-based selection \citep{axiotis_data_2024}. On the Gaussian pool,
where the classes overlap, it does not lower the impurity, and its effect on
the test error is small and of either sign. The distance sets the size of
these effects, which shrink once the representation carries the labels.

Our experiments have several limitations. Except for the network of
Appendix~\ref{sec:exp-retrain}, they use linear models on fixed representations
and a single selection round. The retrained network is small, trained by
L-BFGS on at most \LabelsMax{} points, and the pools are small; deeper
networks trained by stochastic gradient on larger pools, the setting in
which core-set selection was introduced, may behave differently. \rev{The
scales of Prop.~\ref{prop:scale} also depend on the size of the pool. When
points are added to it, $\Delta_d$ cannot increase and, for a given labelled
sample, the radius cannot decrease, whereas the mean distance can change in
either direction.} The regularisation is not rescaled
across representations, which affects the cosine distance most. The
Lipschitz constant that we use is valid over the whole representation
space; a constant restricted to the pairs formed by a pool point and its
nearest labelled point would be smaller, and we do not report it. Labels
are treated as deterministic on the image pools. \rev{We did not compute the
bounds for the $0$--$1$ loss, for which our experiments only evaluate the
ranking of the selections. For its gap,} $m(r_d,\cE)$ is below $1/2$ at
every budget on \texttt{digits} and above $1/2$ at every budget on the
Gaussian pool, as for the loss $1-p_h$, and on MNIST pixels its interval
lies above $1/2$ from $b=\BudgetLargeFrom$ on; in the learned feature space
it stays between \ZoRgFeatMin{} and \ZoRgFeatMax{} with no interval above
$1/2$, so that our conclusion in this space depends on the loss. The
misalignment depends on the family of selections that are compared, and
ours is small; its intervals are not corrected for the number of
comparisons. Proposition~\ref{prop:scale} and Corollary~\ref{cor:budget},
however, depend only on the geometry of the labelled pool and can be
evaluated on any benchmark.

\section{Conclusion}
\label{sec:conclusion}

We read the proof of the bound that motivates core-set active learning
before its statement. Written with the assumption that its proof uses, a
zero conditional risk at the labelled points, the bound holds for any
dissimilarity and any selection rule that uses no label outside the sample,
and under deterministic labels the proof also gives a bound in mean distance
without the factor $C$ and without zero-loss assumption. On our image pools,
neither bound is informative in any run, with fixed or retrained
representations, and the radius does not rank selections consistently with
the gap. Minimising the mean distance rather than the largest one improves
the purity and the accuracy over random sampling on these pools, while on
the Gaussian pool, whose classes overlap, it does not lower the impurity and
changes the test error little. We suggest that guarantees used as design
principles be reported with the scale at which they can be informative and
with their misalignment, and that the distance be treated as a part of the
guarantee.

\bibliographystyle{plainnat}
\bibliography{refs}

\appendix
\section{Role of assumption (H4)}
\label{app:h4}

When labels are random at the labelled points, a zero training loss does
not suffice for \eqref{eq:sener}. Consider any pool, two classes with
$\eta\equiv(\tfrac12,\tfrac12)$, a single labelled point $s$ and the
$0$--$1$ loss. The model class consists of the constant classifiers, and
training on $s$ retains the one that predicts $y_s$ everywhere. For every
classifier of this class the loss does not depend on $\bx$, and $\eta$ is
constant, so that the Lipschitz assumptions of \eqref{eq:sener} hold with
$\lambda^\ell=\lambda^\eta=0$ for every model of the class, and the
training loss is zero. The right-hand side of \eqref{eq:sener} reduces to
$\sqrt{\log(1/\gamma)/(2N)}$, whereas the gap is the proportion of pool
points whose label differs from $y_s$, which concentrates around $1/2$.
Inequality \eqref{eq:sener} thus fails with a probability that tends to one
as $N$ grows. Assumption (H4) does not hold in this example, since the loss
of the other class is one at $\bx_s$ while its probability is $1/2$, and the
proof of \citet{sener_active_2018}, which uses (H4), is not affected by it.

\section{Experimental details}
\label{app:details}

The cosine distance $\sqrt{2(1-\cos)}$ is the Euclidean distance between
pixels normalised to unit norm, and it is a metric. It is an increasing
function of the cosine dissimilarity $1-\cos$, which is not a metric. Both
give the same nearest labelled points, hence the same impurity, and the same
$k$-center traversal; the mean distances and the $k$-medoids selection
differ, and all our quantities are computed with the cosine distance. The
learned representations are the $128$ hidden activations of a
one-hidden-layer network trained on $10^2$, $10^3$, $5\cdot10^3$ or
$2\cdot10^4$ labelled images, with a test accuracy of \AccA, \AccB, \AccC{}
and \AccD.

Each pool comes with a held-out test sample. The initial sample is common to
all the distances of a pool, the two greedy traversals are initialised with
it, and the logistic regression has the same regularisation for all the
representations. The constant $\lambda^\ell$ is valid over the whole
representation space. \rev{The $0$--$1$ loss is not Lipschitz.
On a pool of distinct points it satisfies (H2) with the constant
$L/\min_{i\notin\bS}d_i$, but we did not compute the bounds for it, and our
experiments with this loss only evaluate the ranking of the selections.} Means and paired
differences come with $95\,\%$ Student intervals, misalignments with
$95\,\%$ Wilson intervals whose sample size is the number of repetitions.
These intervals are pointwise, and they describe the variability of the
initial sample and of the random draws for a fixed pool.

\section{Additional results}
\label{app:results}

This appendix gives the misalignments of the quantities of
Sect.~\ref{sec:theory}, their behaviour along the distances of the MNIST
pool and with a network retrained at each cycle, and the terms of the cell
bound.

\subsection{Misalignment}
\label{app:mis}

\begin{figure}[!htb]
  \centering
  \includegraphics[width=\textwidth]{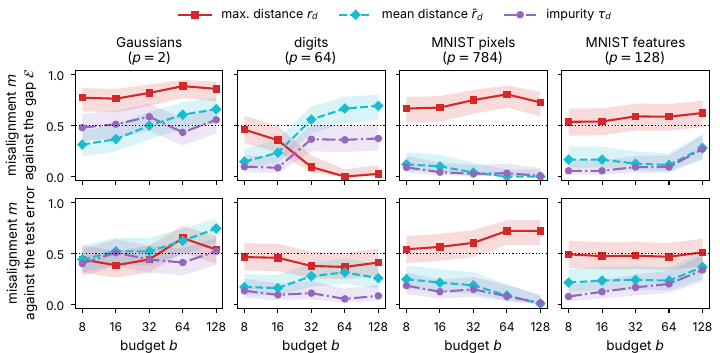}
  \caption{Misalignment (Definition~\ref{def:mis}) of the maximal distance,
  the mean distance and the impurity with the gap (top) and with the test
  error (bottom), over the three pairs of strategies. Shaded: $95\,\%$ Wilson
  interval; dotted line: $m=1/2$.}
  \label{fig:misalignment}
\end{figure}

Figure~\ref{fig:misalignment} compares the three strategies two by two,
a value below $1/2$ meaning that the quantity ranks the selections as the
gap or the test error does. For CS against random sampling alone, the radius
and the mean distance order the selections oppositely in every repetition on
the MNIST pool at $b=\LadderBudget$, under every distance, so that only a
third strategy can separate them.

\subsection{Choice of distance}
\label{sec:exp-dist}

\begin{figure}[!htb]
  \centering
  \includegraphics[width=\textwidth]{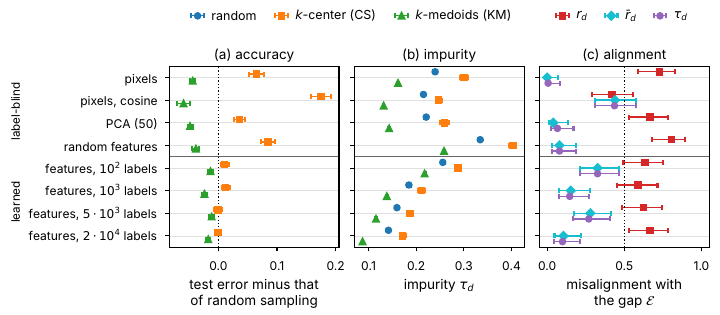}
  \caption{MNIST pool under \NDist{} distances, $b=\LadderBudget$
  (\RepsLadder{} repetitions, $95\,\%$ intervals). (a) Paired difference of
  test error with random sampling. (b) Impurity of each selection.
  (c) Misalignment with the gap over the three pairs of strategies.}
  \label{fig:distances}
\end{figure}

Figure~\ref{fig:distances} follows the MNIST pool along its \NDist{}
distances at $b=\LadderBudget$. CS is less accurate than random sampling
under the label-blind distances, most of all under the cosine distance, for
which the common regularisation makes the model underfit, and under which no
misalignment can be told from $1/2$. The impurity of random sampling is not
monotone along the ladder, and it falls below its value under every
label-blind distance only with features learned on $10^3$ labels or more.

\subsection{Retrained representation}
\label{sec:exp-retrain}

\begin{table}[t]
  \centering
  \caption{Fixed representation and network retrained at each cycle,
  $b=\BudgetMax$ (means over \RepsMain{} repetitions). Fixed: Euclidean
  distance of Sect.~\ref{sec:exp-protocol}, on the pixels for MNIST.
  Paired difference of test error with random sampling ($95\,\%$ interval),
  and misalignment of the maximal distance and of the impurity with the gap,
  CS against random sampling.}
  \label{tab:retrain}
  \small\setlength{\tabcolsep}{3.5pt}
  \begin{tabular}{lrrrrrrrr}
    \toprule
     & \multicolumn{2}{c}{CS $-$ random} & \multicolumn{2}{c}{KM $-$ random} & \multicolumn{2}{c}{$m(r_d,\cE)$} & \multicolumn{2}{c}{$m(\tau_d,\cE)$} \\
    \cmidrule(lr){2-3}\cmidrule(lr){4-5}\cmidrule(lr){6-7}\cmidrule(lr){8-9}
    Pool & fixed & retrained & fixed & retrained & fixed & retr. & fixed & retr. \\
    \midrule
    \TableRetrainRows
    \bottomrule
  \end{tabular}
\end{table}

The network has one hidden layer of \RtHidden{} ReLU units, the architecture
of our feature extractors. It is trained from scratch on the labelled sample,
with the $L^2$ penalty of the logistic regression, at each of five cycles
that add \RtCycles{} points; the initial samples and the batches of random
sampling are those of the fixed representations, and the network is trained
on the pixels for MNIST. The batch of a cycle is selected with the Euclidean
distance in the hidden layer of the network of the previous cycle, that of
$\bZ_0$ for the first one. The quantities of Sect.~\ref{sec:theory} are
measured in the hidden layer of the network trained on the resulting sample,
in which its output is a softmax-linear model, so that $\lambda^\ell$ is
computed as before. This distance depends on the labels of the sample, and
the statements of Sect.~\ref{sec:theory} still apply to it
(Sect.~\ref{sec:statement}).

Table~\ref{tab:retrain} compares the two settings at the largest budget.
The conclusions of the fixed distances hold on the Gaussian pool and on
\texttt{digits} but not on MNIST, where CS becomes more accurate than random
sampling at every budget and \rev{no consistent ordering of the gap by the
radius is detected, every interval of $m(r_d,\cE)$ containing $1/2$}.
In every run on the image pools, $\rbar_d/\Delta_d\ge\RtRbarDeltaMin$, so
that the bounds that do not use the labels of the pool are not informative,
whereas the cell bound on the pool loss falls below $L$ in \RtCellDig{} of
the \RtRunsPool{} runs on \texttt{digits} and in \RtCellMn{} on MNIST.

\subsection{Terms of the cell bound}
\label{sec:exp-terms}

\begin{table}[t]
  \centering
  \caption{Terms of the cell bound \eqref{eq:pointwise} for the loss
  $1-p_h(y\mid\bx)$ ($L=1$) on the MNIST pool, CS at $b=\LadderBudget$
  (means over \RepsLadder{} repetitions): loss on the pool, cell-weighted
  loss on the sample, left-hand side, the two terms of the right-hand side and
  their sum.}
  \label{tab:bound}
  \small\setlength{\tabcolsep}{6pt}
  \begin{tabular}{lrrrrrr}
    \toprule
    Distance & $\Rh_{\bX}$ & $\Rh^{\nn}_{\bS}$ & $|\Rh_{\bX}-\Rh^{\nn}_{\bS}|$ & $\lambda^\ell\rpure_d$ & $\tau_d$ & $\lambda^\ell\rpure_d+\tau_d$ \\
    \midrule
    \TableBoundRows
    \bottomrule
  \end{tabular}
\end{table}

Table~\ref{tab:bound} evaluates the cell bound, the tightest of
Sect.~\ref{sec:theory}, which holds in every run as it must. Under
\NLipAbove{} of the \NDist{} distances, the Lipschitz term alone exceeds the
range $L=1$ of the loss; the exception is the cosine distance, under which
the model underfits and is nearly flat. As a bound on
$|\Rh_{\bX}-\Rh^{\nn}_{\bS}|$, the cell bound is informative in
\NCosRhsBelow{} of the \NRunsPair{} runs under the cosine distance, in every
run on the Gaussian pool, and in no run of the other pool--distance pairs. As
a bound on the pool loss, its right-hand side
$\Rh^{\nn}_{\bS}+\lambda^\ell\rpure_d+L\tau_d$ is never below \CellMin{}
over the \NRunsDet{} runs with deterministic labels, and that of
\eqref{eq:cfree} is at least as large. The right-hand side of
\eqref{eq:sener}, which bounds the gap, is at least $C\,r_d/\Delta_d>C$ in
every run.

\end{document}